\documentclass[
aip,
amsmath,
amssymb,
 reprint
]{revtex4-1}
\usepackage{algorithm}
\usepackage{algpseudocode}
\usepackage{amsmath}
\usepackage{amssymb}
\usepackage{graphicx}
\usepackage{graphicx}
\usepackage{dcolumn}
\usepackage{bm}

\usepackage[utf8]{inputenc}
\usepackage[T1]{fontenc}
\usepackage{mathptmx}
\usepackage{etoolbox}
\usepackage{xcolor}

\makeatletter
\def\@email#1#2{%
 \endgroup
 \patchcmd{\titleblock@produce}
  {\frontmatter@RRAPformat}
  {\frontmatter@RRAPformat{\produce@RRAP{*#1\href{mailto:#2}{#2}}}\frontmatter@RRAPformat}
  {}{}
}%
\makeatother
\begin{document}

\preprint{AIP/123-QED}

\title[Reservoir Computation with Networks of Differentiating Neuron Ring Oscillators]{Reservoir Computation with Networks of Differentiating Neuron \\ Ring Oscillators}
\author{Alexander Yeung}
\email[]{ayeung@umass.edu}
\affiliation{College of Computer \& Information Sciences, University of Massachusetts Amherst}

\author{Peter DelMastro}
\email[]{pdelmastro@vt.edu}
\affiliation{Department of Mathematics, Virginia Tech}
\author{Arjun Karuvally}
\email[]{akaruvally@cs.umass.edu}
\affiliation{College of Computer \& Information Sciences, University of Massachusetts Amherst}
\author{Hava Siegelmann}
\email[]{hava@cs.umass.edu}
\affiliation{College of Computer \& Information Sciences, University of Massachusetts Amherst}
\author{Edward Rietman}
\email[]{erietman@umass.edu}
\affiliation{College of Computer \& Information Sciences, University of Massachusetts Amherst}
\author{Hananel Hazan}
\email[]{Hananel@hazan.org.il}
\affiliation{Allen Discovery Center, Tufts University}

\date{\today}

\begin{abstract}
Reservoir Computing is a machine learning approach that uses the rich repertoire of complex system dynamics for function approximation. Current approaches to reservoir computing use a network of coupled integrating neurons that require a steady current to maintain activity. Here, we introduce a small world graph of differentiating neurons that are active only when there are changes in input as an alternative to integrating neurons as a reservoir computing substrate. We find the coupling strength and network topology that enable these small world networks to function as an effective reservoir. We demonstrate the efficacy of these networks in the MNIST digit recognition task, achieving comparable performance of 90.65\% to existing reservoir computing approaches. The findings suggest that differentiating neurons can be a potential alternative to integrating neurons and can provide a sustainable future alternative for power-hungry AI applications.

\end{abstract}
\maketitle

\begin{quotation}
Reservoir computing is an approach to machine learning whereby the dynamics of a complex system are used in tandem with a simple, often linear, machine learning model for a designated task. While many efforts have previously focused their attention on integrating neurons, which produce an output in response to large sustained inputs, we focus on using differentiating neurons, which produce an output in response to large \textit{changes} in input. The dynamics of differentiating neurons naturally give rise to oscillatory dynamics when arranged in rings, where we study their applicability in reservoir computing. We find that in arranging rings of differentiating neurons into small world network topologies, networks of these structures were able to perform comparably to other similar reservoir computing approaches on the MNIST handwritten digit recognition task. 
\end{quotation}


\section{Introduction}
\label{sec:intro}

\begin{figure*}
      \centering
      \includegraphics[width=0.9\textwidth]{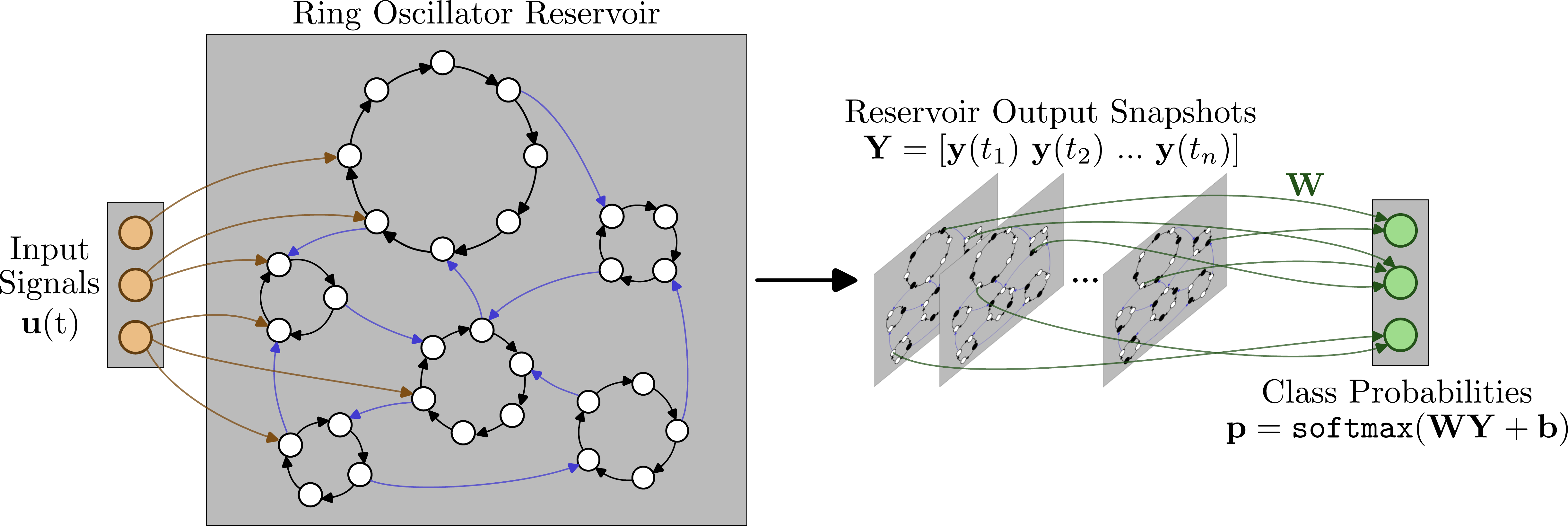}
    \caption{
    Reservoir architecture for computing with networks of differentiating ring oscillators.
    The $j$-th ring oscillator the reservoir is characterized by its time-dependent internal state $\mathbf{v}^{(j)}(t) \in \mathbb{R}^{n_j}$ and output vector $\mathbf{y}^{(j)}(t) \in \{0,1\}^{n_j}$, where $n_j$ denotes the number of differentiating neurons forming that ring. 
    The $R$ ring oscillators in the reservoir are organized into a small world network, resulting in a reservoir with $N = \sum_{j=1}^R n_j$ neurons overall.
    The recurrent weights therefore consist of both \textit{intra}-ring and \textit{inter}-ring connections, depicted as black and blue in the diagram, respectively. 
    The network is driven by external input $\mathbf{u}(t) \in \mathbb{R}^{B}$ for an interval of time $[0,T]$, during which multiple output-snapshots $\mathbf{y}(t_i) = [\mathbf{y}^{(1)}(t_i), ..., \mathbf{y}^{(R)}(t_i)] \in \{0,1\}^N$ are recorded.
    For image classification as presented in this paper, the input signal $\mathbf{u}(t)$ is a sliding window embedding of Hilbert-curve traversal of the image, so that $B$ pixels of the image are being input to the network at each point in time (cf. Sec. \ref{sec:data-and-prediction} for more details).
    The class probability predictions $\mathbf{p} \in [0,1]^{\# \textrm{classes}}$ are obtained by a linear+softmax layer on top of the aggregated snapshot matrix $\mathbf{Y} = [\mathbf{y}(t_1),...,\mathbf{y}(t_n)]$.
    }
    \label{fig:architecture-summary}
\end{figure*}

The current AI revolution is driven by deep neural networks (DNNs) that are powered by fast processing hardware accelerators. As these DNNs grow in size and complexity, yielding improved performance, so too does the computational cost of training and inference \cite{Stiefel2023TheEC}. This motivates the need for alternative models that are capable of similar performance while consuming less energy. One culprit for the high energy demand of DNNs is their fundamental building block - the integrating neuron. Integrating neurons respond to inputs based on a decaying history. They can be modeled by a resistor-capacitor (RC) circuit, where the internal state evolves over time to integrate past inputs, resulting in an output that resembles a smoothed and slightly time-shifted version of the input signal. This behavior arises from the integrating neuron’s tendency to accumulate input over time, producing a large output in response to sustained, gradual changes.


Here, we investigate neural networks of differentiating neurons as an alternative, which unlike integrating neurons respond specifically to rapid changes in the input signal. The output of a differentiating neuron is computed from the time derivative of its internal state, making it sensitive to deviations from the current input average. Thus, differentiating neurons produce large outputs when there are sudden \textit{changes} in the input, such as a switch in signal polarity. This characteristic allows them to act as detectors for rapid transitions rather than cumulative changes. In this way, differentiating neurons provide a fundamentally different dynamic behavior compared to integrating neurons.

The differentiating neural network we propose is organized in two levels. In the first level, the neurons are organized in a ring where the current flows in one direction, naturally giving rise to oscillatory dynamics. In the second level, these rings are organized into small world graphs \cite{Watts1998} and connected via weak coupling between individual neurons in each ring. The second level organization enables the ring oscillators to interact and produce complex dynamical behavior necessary for reservoir computing.

When differentiating neurons are arranged in a ring configuration, they naturally give rise to oscillatory dynamics. In such rings, each neuron is connected to its neighbor in a circular pattern, with the output of each neuron serving as the input to the next. When a pulse of activity propagates through the ring, it triggers successive neurons to fire, maintaining the oscillation, and potentially allowing these pulses of activity to circulate indefinitely. The period and stability of these oscillations depend on various factors, including the number of neurons in the ring and the coupling dynamics. As the pulse travels around the ring, the system effectively encodes a form of dynamic memory in its oscillatory state. This characteristic oscillation is a key property that makes rings of differentiating neurons useful for computational tasks involving temporal data.


Differentiating neurons offer a compelling alternative to integrating neurons for energy-efficient computation, since they activate only in response to changes in input rather than requiring a continuous current. While early applications of small, manually designed networks of differentiating neurons focused on tasks such as robotic control \citep{Still98} and motion tracking \citep{Frigo98}, expanding their use to larger networks unlocks new possibilities for machine learning. By tapping into the potential of differentiating neurons in the realm of reservoir computing, these networks can perform complex tasks such as digit recognition while potentially reducing energy consumption. This makes differentiating neural networks a promising avenue for sustainable AI applications, particularly in resource-constrained environments or low-power hardware implementations.


Previous work on differentiating neural networks required networks to be designed by hand and could only be applied to small-scale problems like the control of robotic insects \cite{Still98} and rudimentary motion tracking \cite{Frigo98}. We apply the principles of Reservoir Computing (RC), an unconventional approach \cite{Jaeger_2021} in machine learning, to develop a differentiating ring oscillator network that can be scaled to solve the MNIST \cite{MNIST} digit recognition task. We also investigate how the number of ring oscillators, their coupling strengths and network topologies affect task performance. We find the ideal settings of these hyperparameters that enable differentiating neural networks to be competitive with existing Reservoir Computing models. 


\section{Related work}
\label{sec:related}
In this section, we provide an overview of the Reservoir Computing paradigm and its reliance on dynamical systems for computation, inspired by the Principle of Computational Equivalence \cite{Wolfram2002}. Following this, we detail Reservoir Computation pipelines and highlight key implementations, including differentiating neuron architectures. Finally, we explore the dynamics of differentiating neurons and their feasibility as a reservoir.

\subsection{Dynamical Systems as Reservoir Computers}

The Principle of Computational Equivalence (PCE), introduced by Stephen Wolfram \cite{Wolfram2002}, posits that a wide range of natural systems can achieve computation in some capacity. This concept broadens the scope of computation to include physical and dynamical systems such as excitable media, programmable matter, and biological substrates \cite{stepney2012, FirstRCPaper}. Reservoir computing makes use of this principle, utilizing dynamical systems as computational substrates by encoding input signals into high-dimensional, nonlinear representations. These reservoirs, whether physical or digital, offer a computational paradigm distinct from traditional logical operations, where a model would instead have to learn to encode inputs into higher dimensions.

Reservoirs vary widely in their physical or simulated realizations. Examples include digital echo state networks \cite{EchoStateMNIST2016}, physical systems such as programmable quantum matter \cite{PZT-Cube}, and hybrid approaches employing high-precision numerical simulations. A notable hybrid example is the skyrmion lattice reservoir simulated via the classical Heisenberg model by \citet{skyrmion_res}. Given its similarity to our approach, we later compare our implementation with \citet{skyrmion_res} to contextualize our results.

RC architectures, first introduced by \citet{FirstRCPaper, jaeger_echostate}, share a common pipeline. An input signal, typically representing time series data, perturbs a reservoir, where the inherent dynamics of the system apply a nonlinear transformation. Then a simple, often linear, machine learning model is trained on the output for downstream tasks. This simplicity in training, leveraging the dynamics of a non-linear system as opposed to backpropogation, is a key distinction between reservoir computing and traditional artificial neural networks.

Dynamical systems near the edge of chaos, as discussed by \citet{LANGTON199012, dambre2012}, are particularly effective reservoirs. Operating at this boundary ensures a balance between stability and responsiveness, maximizing the expressiveness of the system. For instance, networks of oscillatory circuits can optimize performance by maintaining weak coupling and avoiding over-saturation, as demonstrated in \citet{Explosive-Sync}. In our implementation, we tune the hyperparameters of our differentiating ring oscillator networks to achieve optimal dynamics.

\subsection{Differentiating Neuron Architectures}

Differentiating neurons are artificial neurons described by differential equations wherein the output of the neuron is a function of the time derivative of the voltage across its capacitor. First developed by \citet{Hasslacher97}, these neurons have been widely applied across various domains, including satellite control \citep{satellite_ctrl}, robotic control inspired by living organisms \citep{Hasslacher95}, and pattern recognition \citep{Schmitt_trigger_pattern}.

From the perspective of electronic circuits, differentiating neurons are modeled as a resistor-capacitor circuit, where $v$ describes the voltage across the capacitor, $y$ denotes the output voltage, and $u$ denotes the input voltage. Then the internal state of the neuron $v$, modulated by the time-constant $\tau$, is updated according to the following differential equation:

\begin{equation}
\tau \frac{dv(t)}{dt} = u(t) - v(t)
\end{equation}

In contrast to differentiating neurons, traditional neural networks rely on \textit{integrating neurons}, whose outputs represent a decaying integral of past inputs over time. This means that integrating neurons respond primarily to sustained input magnitudes. Differentiating neurons, on the other hand, compute their output as the \textit{time derivative} of their internal state, making them particularly sensitive to rapid changes in input rather than steady-state values. As a result, the outputs of integrating neurons capture long-term trends in their inputs, whereas differentiating neurons' outputs emphasize transient dynamics. In our case this is described as:

\begin{equation}
y(t) = \phi(\tau \dot{v}) = \phi\big(u(t) - v(t)\big)
\end{equation}

In our simulations, we use binary-output differentiating neurons that are modeled as a Schmitt trigger logic gate which enables short-term memory retention in hysterisis \cite{Hasslacher95}. As a result, changes in output for each neuron are parameterized by crossing an upper and lower threshold of the Schmitt trigger, denoted as $v_{thh}$ and $v_{thl}$ respectively. Thus, our output function $y$ is:

\begin{equation}
y_i(t) =  \begin{cases}
0, & \tau\dot{v}_i(t) \geq v_{thh} \\
0, & \tau\dot{v}_i(t) \in [v_{thl}, v_{thh}]) \ \textrm{and} \ y_i(t^{-}) = 0 \\
1, & \textrm{otherwise}
\end{cases}
\end{equation}

Where $y_i(t^{-})$ denotes that the output of neuron $i$ immediately before crossing one of the thresholds. It should be noted that because these neuron architectures make use of an inverter, 0 denotes that the neuron is firing and 1 denotes that it is dormant.

\subsection{Networks of Differentiating Neuron Oscillators as Reservoir Computers}

\begin{figure}
    \centering
    \includegraphics[width = \columnwidth]{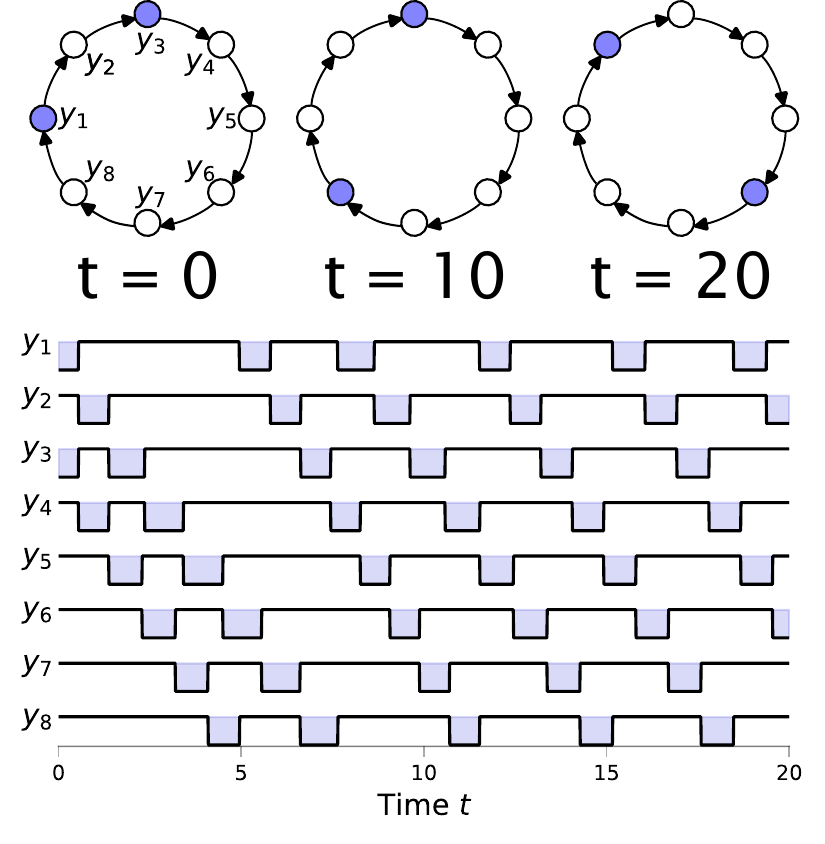}
    \caption{Dynamics of a ring of 8 differeniating neurons. The overall dynamics of the ring oscillator at time of initialization and ten and twenty time steps thereafter are displayed on the top, while the dynamics of each neuron $y_i$ are depicted on the bottom. The shaded regions on both the top and the bottom indicate that a neuron is firing.}
    \label{fig:ring-oscillator}
\end{figure}

\citet{Hasslacher97} first described the dynamics of differentiating neurons composed in ring oscillators while also highlighting some potential applications. We intend to build on this work, using differentiable ring oscillator networks as reservoirs for machine learning tasks. Figure \ref{fig:ring-oscillator} illustrates the structure and dynamics of ring oscillator networks we consider in the paper.

An \( n \)-ring oscillator consists of \( n \) neurons arranged in a ring, where the output of the \( i \)-th neuron serves as the input to the \((i+1)\)-th neuron (modulo \( n \)). The system's behavior is governed by the differential equations:
\[
\tau_i \frac{d v_i}{dt} = y_{i-1}(t) - v_i(t), \quad i = 1, 2, \dots, n,
\]
where \( y_i(t) = \phi(\tau_i \dot{v}_i) \) is the output of neuron \( i \), and \( i-1 \) is considered modulo \( n \). With a Schmitt trigger as the activation function \( \phi \), the neurons can transmit pulses that persist and circulate indefinitely around the ring \cite{Hasslacher97}. 

To extend this mechanism into a reservoir computing architecture, we investigated how to interconnect individual ring oscillators into a larger network. Prior work such as \citet{DelMastro2025TransientDI},
found that a grid of ring oscillators, saturated too quickly for pattern recognition, so we explored a small-world with rings of random sizes to be nodes in a small-world network\cite{Watts1998}. The transition from a reservoir of oscillators with a grid topology to a small-world topology substantially enhances the computational capabilities of the differentiating neural network. In a grid topology, information propagation is constrained by the nearest-neighbor coupling, resulting in a finite velocity of information transfer through the system. This configuration limits the mixing of information across distant spatial regions of the reservoir, requiring many time steps before signals can interact across the full domain. Consequently, the system's memory capacity and ability to learn complex temporal features from the input signals are restricted by this rigid communication structure.

Small-world topologies overcome these limitations by introducing strategic, long-range connections that maintain most of the local coupling structure while adding critical shortcuts across the domain. These shortcuts reduce the effective path length between any two points in the reservoir, enabling rapid information mixing throughout the system without requiring a proportional increase in physical size or number of oscillators. The resulting architecture preserves the energy conservation constraints while significantly expanding its expressivity, allowing it to learn more complex temporal dependencies and improving classification accuracy on tasks such as digit recognition.
\section{Materials \& Methods}
\label{sec:methods}
 The network we consider is composed of \( n_{\text{rings}} \) oscillatory rings, with each ring containing a random number of neurons drawn uniformly from over the range of three to ten neurons per ring. Three is the minimum number of neurons we consider to form a ring, since rings with two or less neurons may not circulate their pulses indefinitely.  Neurons within a ring interact through intra-ring connections, while inter-ring interactions occur via small-world coupling, modeled using the Watts-Strogatz graph construction \cite{Watts1998}. This method introduces a degree of randomness in connectivity, governed by a rewiring probability \(p\). Additionally, the strength of inter-ring couplings is controlled by a parameter \( \epsilon \) that modulates the influence of one ring on another. 

A subset of rings is reserved for input and is externally stimulated by a sequence of inputs with frequency controlled by the connectivity parameter \(p\) above.

Thus, the input voltage \( u_i^{(j)} \) for neuron \( i \) in the $j$-th ring is given by:
\[
u_i^{(j)} = y^{(j)}_{i-1 \ \text{mod} \ n} + \epsilon \sum_{k \ \in \ N(i,j)} y^{(k)}_{i-1 \ \text{mod} \ n} + \sum_{\ell \ \in \ I(i,j)} u^{(\textrm{ext})}_\ell
\]
where:
\begin{itemize}
    \item \( y^{(j)}_{i-1 \ \text{mod} \ n} \) accounts for the \textit{intra}-ring input contribution received from the neuron $(i-1)$-th neuron in ring $j$,
    \item \( N(i,j) \) denotes the set of external rings connected to neuron \( i \) in ring $j$ via the small-world coupling,
    \item \( I(i,j) \) represents the set of external signal(s) to neuron $i$ in oscillator $j$. This set may be empty if the neuron is not chosen to receive input.
\end{itemize}

Note that since every oscillator in our network must contain at least three neurons, we arbitrarily select a first and second neuron in each oscillator to send and receive outputs to external neurons in other oscillators modulated by a coupling constant, as denoted by $N(i,j)$ above.

\begin{figure}
      \centering
      \includegraphics[width = \columnwidth]{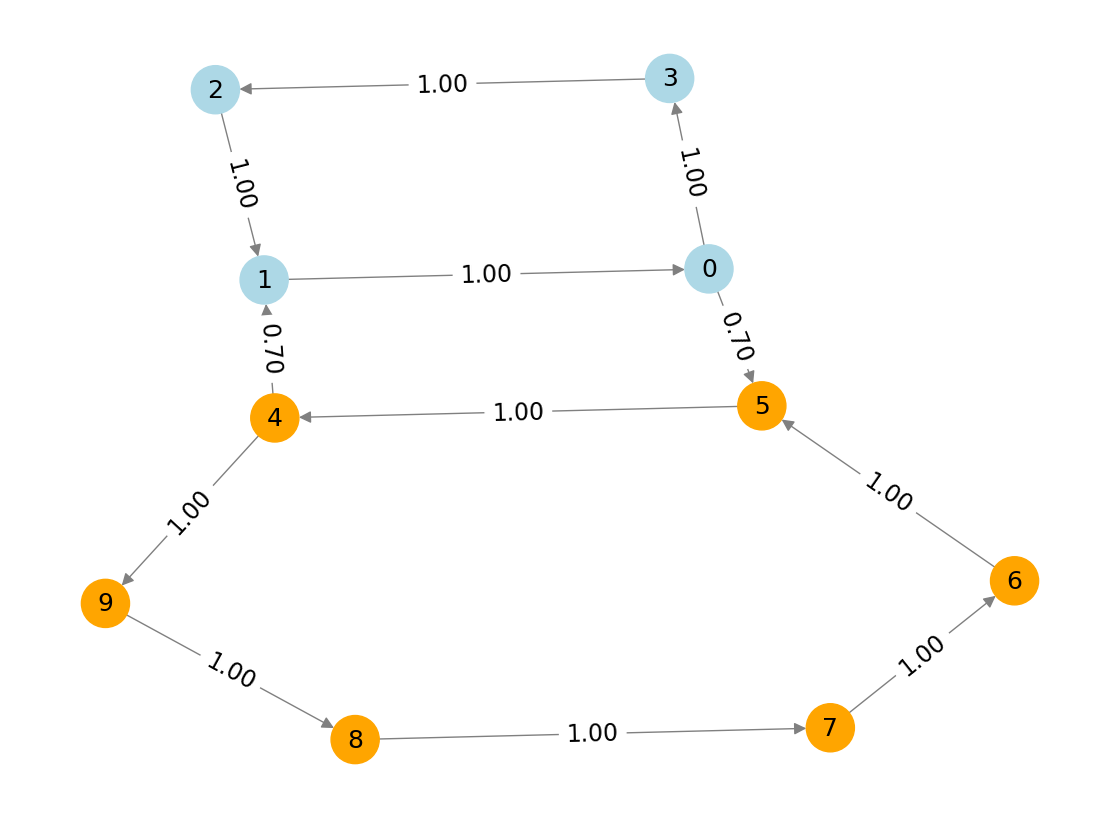}
    \caption{Example coupling of two oscillators, in this case one with four neurons and one with six. The edges of the graph are labeled with the strength of their connectivity. Here, the coupling constant $\varepsilon$ is chosen to be 0.7.}
    \label{fig:two-oscillators}

\end{figure} 

\begin{figure}
      \centering
      \includegraphics[width = \columnwidth]{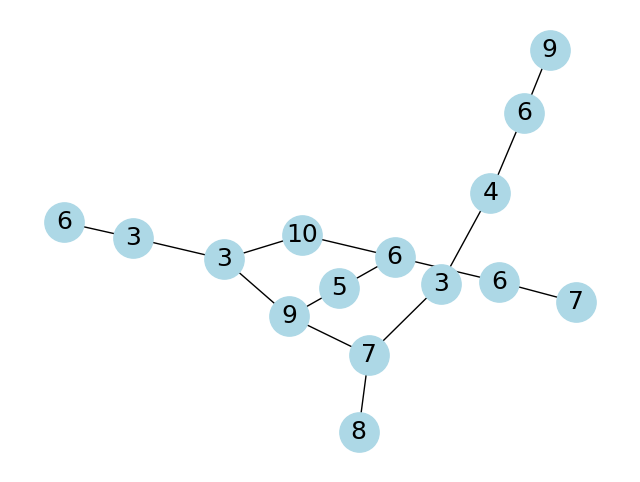}
    \caption{Example small-world network constructed via the Watts-Strogatz method. The individual oscillators are labeled according to the number of differentiating neurons they contain. The connections are coupled as shown in Figure~\ref{fig:two-oscillators}.}
    \label{fig:small-world-network}
\end{figure}

Our networks exhibit organization on two levels. On the more granular level, oscillators are connected through the coupling mechanisms described above. Figure \ref{fig:two-oscillators} shows how two rings are coupled to each other, where one of the rings has four differentiating neurons and the other has six differentiating neurons. On a broader level, we determine which oscillators are coupled to one another by arranging our network in a small-world topology, generated by the Watts-Strogatz method \cite{Watts1998}. Figure \ref{fig:small-world-network} depicts a network that exemplifies this overall structure.

\subsection{Initialization and Simulation}

The network is initialized by assigning random states to neurons, with a small number randomly selected to fire at the start. This then begins a warmup phase where the initial dynamics of each ring are simulated independently of one another for a random duration to stabilize oscillations and ensure coherent dynamics before coupling.

Following the warmup phase, the state of each neuron is updated at discrete timesteps, incorporating contributions from intra-ring, inter-ring, and external inputs. We chose time constant $\tau$ for all simulations to be equal to one since we were primarily interested in exploring the role of network topologies and coupling strengths in our networks. 



The role of coupling in shaping network behavior is analyzed in Section~\ref{sec:results}, while implementation details of our simulation, time-series preprocessing, and network construction are provided in Appendix A.

\subsection{Data and Prediction}
\label{sec:data-and-prediction}
\begin{figure}
    \centering
    \includegraphics[width = \columnwidth]{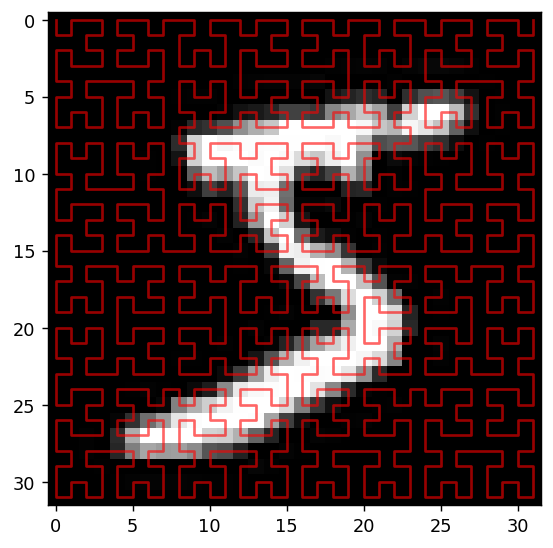}
    \caption{MNIST digit resized and read along Hilbert curve. The original MNIST digit images are 28x28, we resize them to 32x32 to fit the Hilbert curve. Following this transformation, we then read the image along a sliding window as the time-series input to our reservoir.}
    \label{fig:hilbertized}
\end{figure}

To train and evaluate the reservoir, we adapted the MNIST dataset by converting static images into time-series data. Each $28 \times 28$ pixel image is first resized to $32 \times 32$ pixels using bilinear interpolation for compatibility with the network. The pixel values are normalized and augmented with Gaussian noise to enhance robustness. Next, the image is traversed sequentially along a Hilbert curve, a space-filling curve that preserves spatial locality. The traversal outputs a one-dimensional sequence of pixel intensities, maintaining the spatial relationships between neighboring pixels (Figure~\ref{fig:hilbertized}).

To further enhance temporal structure, the Hilbert-ordered sequence undergoes a sliding window embedding. A fixed-size window, determined by the number of input neurons ($n_{\text{in}}$), is applied to extract overlapping subsequences from the traversal. These subsequences form the initial temporal inputs to the reservoir. To match the desired length of the time-series data ($n_{\text{ts}}$), each subsequence is repeated proportionally, with any remainder filled by duplicating entries from the start of the sequence. This process ensures that the input to the reservoir is both length-consistent and spatially meaningful.

The reservoir output is classified by applying a softmax layer on top of a linear transformation that maps the output data to each digit (0--9). The same linear transformation that was learned during training time is used to predict unseen digits during inference.

\section{Results}
\label{sec:results}

\begin{figure}
      \centering
      \includegraphics[width = \columnwidth]{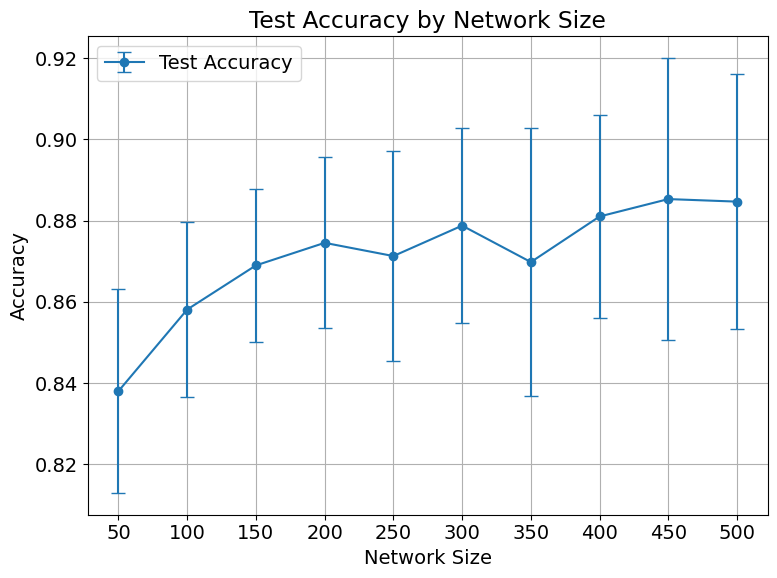}
    \caption{Test accuracy of ring oscillator networks ranging from 50 to 500 rings. All networks were initialized with the same hyperparameters $\varepsilon$ and $p$ both set at 0.5.  The points depict the mean of the 10 trials of each network size, and the error bars display the standard deviation of the trials.}
    \label{fig:accuracy_diffsizes}
\end{figure}

The performance of the reservoir was analyzed as a function of network size and hyperparameter configurations- the coupling constant ($\varepsilon$) and the connectivity constant ($p$). These analyses provide insights into the trade-offs between model complexity, generalization, and overfitting.

\subsection{Hyperparameter Sensitivity}

To assess the robustness of our reservoir architecture to network-level design choices, we conducted a grid search over the coupling strength $\varepsilon$ and connectivity parameter $p$, each varied from 0.1 to 0.9 in increments of 0.1. Each configuration was evaluated across five randomly initialized trials using networks of 300 ring oscillators.

Training accuracy exhibited considerable variability across the hyperparameter grid, with no single region emerging as consistently optimal. This variability suggests that convergence during training may be influenced by factors beyond $\varepsilon$ and $p$ alone, such as initial conditions or stochastic noise in the readout training pipeline. These findings underscore the importance of robustness to hyperparameter variation in practical reservoir computing applications.

In contrast, test accuracy showed more structured behavior. Performance was generally highest when the coupling strength $\varepsilon$ was in the lower to intermediate range (e.g., 0.2--0.4), and results were relatively stable across a broad range of $p$. This suggests that generalization performance is more sensitive to coupling dynamics than to global connectivity in the small-world regime. Notably, the best-performing configuration achieved a test accuracy of 90.65\% for $\varepsilon = 0.2$ and $p = 0.4$, with a corresponding validation accuracy of 88.47\%.

These results support the hypothesis that optimal performance emerges near the \emph{edge of chaos}---a dynamical regime in which the reservoir balances high sensitivity to input perturbations with sufficient stability to avoid signal degradation \citep{LANGTON199012, dambre2012}. Excessively high coupling or dense connectivity tends to push the system toward oversaturation, limiting its ability to generalize to unseen data. By contrast, operating in an intermediate regime allows the system to exhibit rich but tractable dynamics, thereby enhancing its expressiveness.

Together, these findings affirm the value of small-world reservoir topologies with well-tuned coupling parameters, as they provide a flexible substrate for modeling temporally extended input patterns while remaining robust to network variation.

\subsection{Impact of Network Size}

Figure~\ref{fig:accuracy_diffsizes} illustrates the relationship between network size and performance, with all networks using identical hyperparameters. These values were chosen as an intermediate setting to ensure both hyperparameters had equal influence across varying ring sizes. Interestingly, test accuracy began showing diminishing returns at approximately 300 ring oscillators, after which it did not increase at a significant rate. At 300 ring oscillators, the mean test accuracy was 87.9\%, whereas the highest overall test accuracy occurred at a network size of 450 with a mean test set performance of 88.5\%. This stagnation in test accuracy performance suggests potential saturation as network size grows beyond this threshold.

The error bars in Figure~\ref{fig:accuracy_diffsizes}, which represent the standard deviation across all trials show the variability in performance. The increase in variability for larger network sizes indicate that the test performance of larger networks may be more sensitive to initialization or specific data characteristics.


\section{Discussion and Future Works}
\label{sec:discussion}

Our work demonstrates that differentiating neuron oscillators can function as effective reservoir computers. By leveraging small-world topologies and weak coupling strengths, we achieved competitive performance, notably up to 90.65\% accuracy on MNIST digit recognition. This puts our system on par with other in-material (\textit{in materio}) reservoirs, such as the PZT cubes studied by \citet{PZT-Cube} (80\%), diffusive memristors implemented by \citet{memristors} (83\%), and magnetic skyrmion reservoirs described by \citet{skyrmion_res} (90\%).

The analysis of our reservoir revealed key insights on the dynamics of coupled oscillators and their influence on reservoir computing. The ideal network configurations found from our experiments align with prior work by \citet{KuramotoModel}, that found performance peaks near the "edge of chaos," where oscillatory behavior balances between order and disorder. The observed decline in accuracy at higher coupling thresholds suggests that excessive homogeneity may diminish diversity. Another perspective on the coupling-performance relationship is based on the spatial extend enabled by weak couplings. Recent studies \cite{Budzinski2022AnalyticalPO, Benigno2023WavesTO} have shown that spatial organization enables structured memory storage in the propagation of neural activity waves. These propagating waves enable the storage of recent history \cite{Keller2023TravelingWE} and encode temporal information \cite{pmlr-v235-karuvally24a}. We suggest that understanding the relationship between this form of spatial computation and the efficacy of information processing is necessary to better design reservoir computing systems. This relationship remains underexplored and warrants further investigation. 

Additionally, our investigation into network size revealed a plateauing effect in test performance beyond 300 oscillators. This may indicate that additional network complexity does not translate to meaningful improvements in generalization. Instead, it highlights the importance of architectural efficiency and the risk of diminishing returns with increasing network size. The variability observed across different hyperparameter settings further underscores the importance of tuning network dynamics to avoid over-saturation and ensure optimal performance.

\subsection{Future Directions}

Building on these findings, we identify several avenues for future research:  

\begin{enumerate}
    \item \textbf{Application to New Tasks:} Extending this reservoir architecture to time-series data, such as spoken MNIST digits and predictions of physical systems like Lorenz attractors, may reveal its broader utility in machine learning tasks.
    \item \textbf{Network Properties and Metrics:} Investigating the role of network properties, such as homogeneity, synchronization, and topology, may illuminate the conditions that optimize reservoir performance \cite{HAZAN20121597}. Metrics quantifying information diversity or experiments with alternative topologies (e.g., Erdős–Rényi graphs) could refine our understanding.
    \item \textbf{Hardware Implementations:} Translating this architecture into physical systems opens pathways for energy-efficient computation inspired by natural dynamical systems, such as central pattern generators.
\end{enumerate}


\section{Conclusion}

This study demonstrates that networks of differentiating neuron oscillators can function as effective reservoir computers when structured with small-world connectivity and weak coupling strengths. Our experiments revealed that test accuracy stagnates beyond a network size of 300 oscillators, suggesting that larger networks do not necessarily yield better generalization. Furthermore, we observed that optimal performance emerges at intermediate coupling and connectivity values, aligning with the "edge of chaos" hypothesis, where reservoirs balance stability and expressiveness.

These findings contribute to a growing understanding of how network dynamics influence reservoir computing. Future research should explore how these principles extend to tasks beyond static image classification, such as time-series forecasting and control applications. Additionally, investigating the role of network homogeneity, synchronization, and alternative topologies could provide deeper insights into optimizing reservoir architectures. The potential for hardware implementations also presents an exciting avenue for leveraging the efficiency of physical dynamical systems for computation.

By demonstrating competitive performance with other in-material reservoirs, our results reinforce the viability of oscillatory networks for neuromorphic computing. As research in this area progresses, refining the interplay between network structure, coupling dynamics, and task-specific requirements will be key to unlocking the full potential of reservoir computing.

\section*{Author Contributions}

Conceptualization, A.Y., P.D., H.H., E.R.; 
Methodology, A.Y., P.D., H.H., E.R.; 
Software, A.Y., A.K.; 
Validation, A.Y., P.D., A.K., H.H., E.R.; 
Formal Analysis, A.Y., A.K.; 
Investigation, A.Y., A.K.; 
Resources, E.R.; 
Data Curation, A.Y., A.K.; 
Writing – Original Draft Preparation, A.Y.; 
Writing – Review \& Editing, A.Y., P.D., A.K., H.S., H.H., E.R.; 
Visualization, A.Y., P.D.; 
Supervision, H.H., E.R.; 
Project Administration, E.R.

\section*{Data Availability}

The raw data supporting the conclusions of this article will be made available by the authors on request.

\begin{acknowledgments}
We appreciate the assistance of Mohsin Shah for his early work in exploring the dynamics of the differentiating ring oscillator networks used in this work.
\end{acknowledgments}

\nocite{*}
\section{References}
\bibliography{references}

\vfill
\raggedbottom

\appendix
\label{appendix:reservoir_sim}
\section{Reservoir Simulation}

\begin{algorithm}[H]
\caption{Reservoir Computing Simulation}
\label{alg:reservoir_sim}
\begin{algorithmic}[1]
\State \textbf{Parameters:}
\State \quad $n_{\text{rings}}$: Number of ring oscillators
\State \quad $\text{lower}$, $\text{upper}$: Range of ring sizes
\State \quad $\epsilon$: Coupling constant
\State \quad $p$: Adjacency constant
\State \quad $n_{\text{in}}$: Number of inputs
\State \quad $dt$: Length of each time step
\State \quad $T$: Simulation time
\State \quad $v_{\text{thh}}$: Upper voltage threshold
\State \quad $v_{\text{thl}}$: Lower voltage threshold
\State \quad $v_{\text{in}}$: Input as time series data

\State \textbf{Outputs:}
\State $v_{\text{out}}$: History of neuron outputs where $v_{\text{out}}[i,j]$ is the output at time step $i$ of neuron $j$
\State $v_{\text{cap}}$: History of the neuron capacitor values where $v_{\text{cap}}[i,j]$ is the voltage at the capacitor at time step $i$ of neuron $j$

\State \textbf{Initialize Network:}
\State Generate ring sizes $R_i$ uniformly over [lower, upper].
\State $n_{\text{neurons}} \gets \sum_i R_i$
\State $\text{idx} \gets$ indices of the first neuron in each ring.
\State Create $W$ as an adjacency matrix for intra- and inter-ring connections using the Watts-Strogatz model, scaled by $\epsilon$.
\State Create $W_{\text{in}}$ for input connections using Bernoulli trials with parameter $p$, scaled by $\epsilon$.
\State Initialize $v_{\text{cap}}[0] \gets 0.9$ and $v_{\text{out}}[0] \gets 1$.
\State Randomly select initial firing nodes in each ring.
\State Simulate each ring independently for a random duration.

\Function{SchmittTrigger}{$o, v$}
    \If{$v \leq v_{\text{thl}}$} \Return $0$
    \ElsIf{$v \geq v_{\text{thh}}$} \Return $1$
    \Else \Return $o$
    \EndIf
\EndFunction

\State \textbf{Simulate Network:}
\For{$t = 1$ to $\frac{T}{dt}$}
    \State $\alpha \gets e^{-{dt}}$
    \State Update capacitor voltages:
    \State $u \gets W \times v_{\text{out}}[t] + W_{\text{in}} \times v_{\text{in}}[t]$
    \State $v_{\text{cap}}[t+1] \gets \alpha \times v_{\text{cap}}[t] + (1 - \alpha) \times u$
    \State Update outputs:
    \State $v_{\text{out}}[t+1] \gets \text{SchmittTrigger}(v_{\text{out}}[t], u - v_{\text{cap}}[t+1])$
\EndFor

\State \textbf{Main Function:}
\State Initialize the network
\State Simulate the network with input $v_{\text{in}}$
\State Return $v_{\text{out}}$, $v_{\text{cap}}$

\end{algorithmic}
\end{algorithm}

\end{document}